\documentclass[10pt,twocolumn,letterpaper]{article}
\usepackage{cvpr}
\usepackage{times}
\usepackage{epsfig}
\usepackage{graphicx}
\usepackage{amsmath}
\usepackage{amssymb}
\usepackage{graphicx}
\usepackage{amsmath}
\usepackage{amssymb}
\usepackage{algorithm}
\usepackage{algorithmic}
\usepackage{booktabs}
\usepackage{subfigure}
\usepackage{stfloats}
\usepackage{nccmath}
\usepackage{bm}
\usepackage{multirow}
\usepackage{multicol}
\usepackage{array}
\usepackage{url}
\usepackage{algorithm}              
\usepackage{algorithmic}              
\usepackage{xcolor} 
\usepackage{graphics} 
\usepackage{epsfig} 
\usepackage{colortbl}
\usepackage{arydshln}
\usepackage{array}
\usepackage{booktabs}
\usepackage{threeparttable}
\usepackage[shortlabels]{enumitem}

\usepackage[pagebackref=true,breaklinks=true,letterpaper=true,colorlinks,bookmarks=false]{hyperref}

\cvprfinalcopy 

\def\cvprPaperID{3633} 

\ifcvprfinal\fi
\begin{document}

\title{Learning Face Age Progression: A Pyramid Architecture of GANs}
\renewcommand{\thefootnote}{\fnsymbol{footnote}}
\setcounter{footnote}{0}
\author{Hongyu Yang$^{1}$~~ Di Huang$^{1}$\footnotemark\ ~~ Yunhong Wang$^{1}$~~Anil K. Jain$^{2}$\\
$^{1}$Beijing Advanced Innovation Center for Big Data and Brain Computing, Beihang University, China\\
$^{2}$Michigan State University, USA\\
{\tt\small \{hongyuyang, dhuang, yhwang\}@buaa.edu.cn~~jain@cse.msu.edu}
}
\maketitle

\newcommand{\tabincell}[2]{\begin{tabular}{@{}#1@{}}#2\end{tabular}}


\begin{abstract}
The two underlying requirements of face age progression, i.e. aging accuracy and identity permanence, are not well studied in the literature. In this paper, we present a novel generative adversarial network based approach. It separately models the constraints for the intrinsic subject-specific characteristics and the age-specific facial changes with respect to the elapsed time, ensuring that the generated faces present desired aging effects while simultaneously keeping personalized properties stable. Further, to generate more lifelike facial details, high-level age-specific features conveyed by the synthesized face are estimated by a pyramidal adversarial discriminator at multiple scales, which simulates the aging effects in a finer manner. The proposed method is applicable to diverse face samples in the presence of variations in pose, expression, makeup, etc., and remarkably vivid aging effects are achieved. Both visual fidelity and quantitative evaluations show that the approach advances the state-of-the-art.
\end{abstract}

\footnotetext{$^{*}$ Corresponding Author}

%
\vspace{-0.55cm}
\section{Introduction}

Age progression is the process of aesthetically rendering a given face image to present the effects of aging. It is often used in entertainment industry and forensics, \textit{e.g.}, forecasting facial appearances of young children when they grow up or generating contemporary photos for missing individuals. 

The intrinsic complexity of physical aging, the interferences caused by other factors (\textit{e.g.}, PIE variations), and shortage of labeled aging data collectively make face age progression a rather difficult problem. The last few years have witnessed significant efforts tackling this issue, where aging accuracy and identity permanence are commonly regarded as the two underlying premises of its success \cite{Suo:Compositional}\cite{Yang:faceAging}\cite{shu2015personalized}\cite{lanitis2008evaluating}. The early attempts were mainly based on the skin's anatomical structure and they mechanically simulated the profile growth and facial muscle changes w.r.t. the elapsed time \cite{todd1980perception}\cite{wu1994plastic}\cite{ramanathan2008modeling}. These methods provided the first insight into face aging synthesis. However, they generally worked in a complex manner, making it difficult to generalize. Data-driven approaches followed, where face age progression was primarily carried out by applying the prototype of aging details to test faces \cite{Kemelmacher:aging}\cite{Suo:Compositional}, or by modeling the dependency between longitudinal facial changes and corresponding ages \cite{Suo:Concatenational}\cite{Wang:tensorAging}\cite{park2010age}. Although obvious signs of aging were synthesized well, their aging functions usually could not formulate the complex aging mechanism accurately enough, limiting the diversity of aging patterns.

\begin{figure}[t] 
\centering
\includegraphics[width=2.9in]{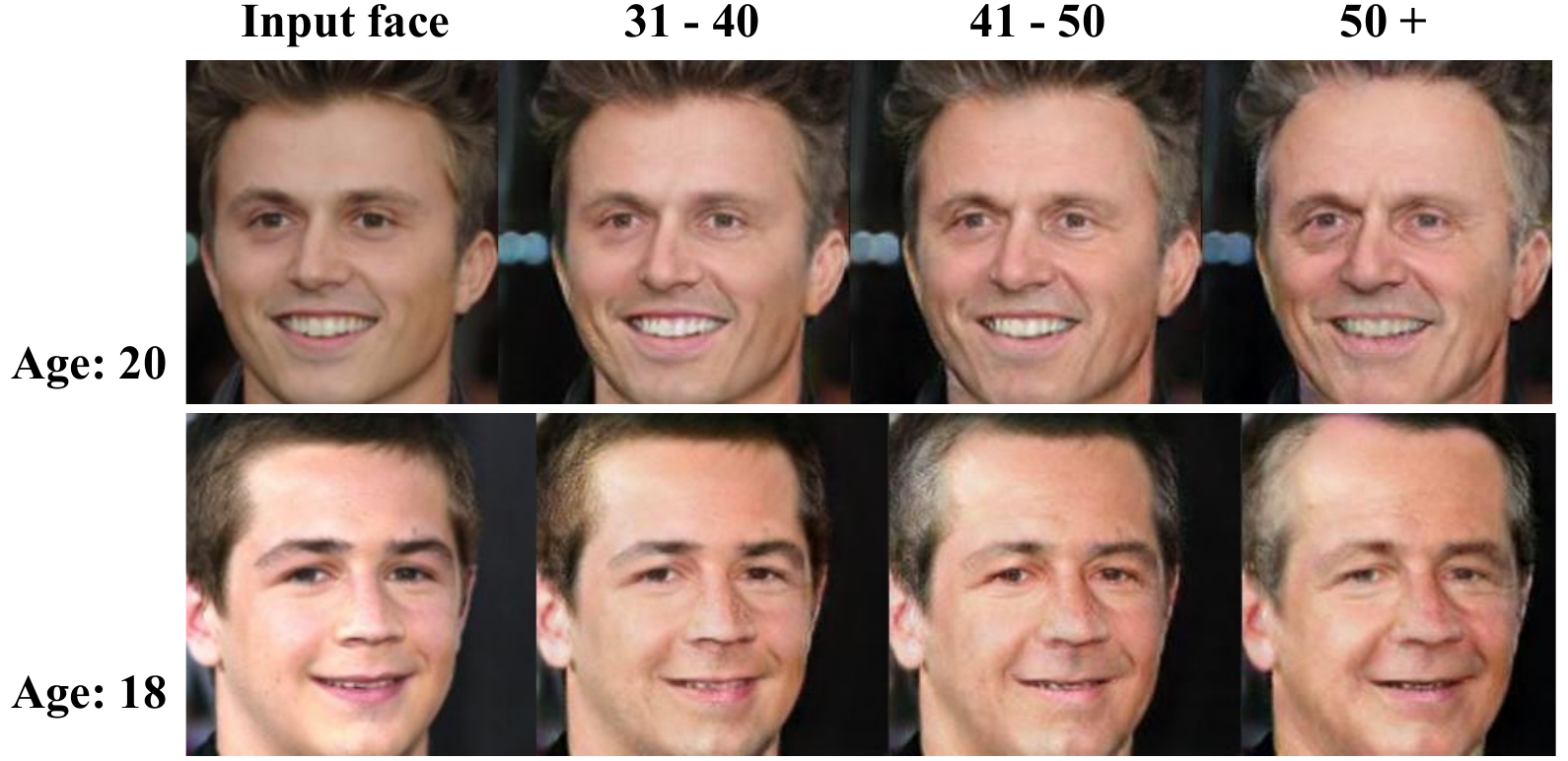}
\caption{Demonstration of our aging simulation results (images in the first column are input faces of two subjects).} 
\vspace{-0.3cm}
\end{figure}

The deep generative networks have exhibited a remarkable capability in image generation \cite{DB16c}\cite{gan}\cite{Phillip2016image2image}\cite{taigman2016unsupervised} and have also been investigated for age progression \cite{Wang:Recurrent}\cite{zhang2017age}\cite{nhan2017temporal}\cite{nhan2016longitudinal}. These approaches render faces with more appealing aging effects and less ghosting artifacts compared to the previous conventional solutions. However, the problem has not been essentially solved. Specifically, these approaches focus more on modeling face transformation between two age groups, where the age factor plays a dominant role while the identity information plays a subordinate role, with the result that aging accuracy and identity permanence can hardly be simultaneously achieved, in particular for long-term age progression \cite{nhan2017temporal}\cite{nhan2016longitudinal}. Furthermore, they mostly require multiple face images of different ages of the same individual at the training stage, involving another intractable issue, \emph{i.e.} intra-individual aging face sequence collection \cite{Wang:Recurrent}\cite{liu2017aging}. Both the aforementioned facts indicate that current deep generative aging methods leave room for improvement.

In this study, we propose a novel approach to face age progression, which integrates the advantage of Generative Adversarial Networks (GAN) in synthesizing visually plausible images with prior domain knowledge in human aging. Compared with existing methods in literature, it is more capable of handling the two critical requirements in age progression, \emph{i.e.} identity permanence and aging accuracy. To be specific, the proposed approach uses a Convolutional Neural Networks (CNN) based generator to learn age transformation, and it separately models different face attributes depending upon their changes over time. The training critic thus incorporates the squared Euclidean loss in the image space, the GAN loss that encourages generated faces to be indistinguishable from the elderly faces in the training set in terms of age, and the identity loss which minimizes the input-output distance by a high-level feature representation embedding personalized characteristics. It ensures that the resulting faces present desired effects of aging while the identity properties remain stable. By estimating the data density of each individual target age cluster, our method does not demand matching face pairs of the same person across two age domains as the majority of the counterpart methods do. Additionally, in contrast to the previous techniques that primarily operate on cropped facial areas (usually excluding foreheads), we emphasize that synthesis of the entire face is important since the parts of forehead and hair also significantly impact the perceived age. To achieve this and further enhance the aging details, our method leverages the intrinsic hierarchy of deep networks, and a discriminator of the pyramid architecture is designed to estimate high-level age-related clues in a fine-grained way. Our approach overcomes the limitations of single age-specific representation and handles age transformation both locally and globally. As a result, more photorealistic imageries are generated (see Fig. 1 for an illustration of aging results).

The main contributions of this study include:

(1) We propose a novel GAN based method for age progression, which incorporates face verification and age estimation techniques, thereby addressing the issues of aging effect generation and identity cue preservation in a coupled manner;
(2) We highlight the importance of the forehead and hair components of a face that are closely related to the perceived age but ignored in other studies; it indeed enhances the synthesized age accuracy;
(3) We set up new validating experiments in addition to existent ones, including commercial face analysis tool based evaluation and insensitivity assessment to the changes in expression, pose, and makeup. Our method is not only shown to be effective but also robust to age progression.

\vspace{-0.15cm}
\section{Related Work}
In the initial explorations of face age progression, physical models were exploited to simulate the aging mechanisms of cranium and facial muscles. Todd \emph{et al.} \cite{todd1980perception} introduced a revised cardioidal-strain transformation where head growth was modeled in a computable geometric procedure. Based on skin's anatomical structure, Wu \textit{et al.} \cite{wu1994plastic} proposed a 3-layered dynamic skin model to simulate wrinkles. Mechanical aging methods were also incorporated by Ramanathan and Chellappa \cite{ramanathan2008modeling} and Suo \emph{et al.} \cite{Suo:Concatenational}. 

\begin{figure*}[t]
\centering 
\includegraphics[width=0.99\textwidth]{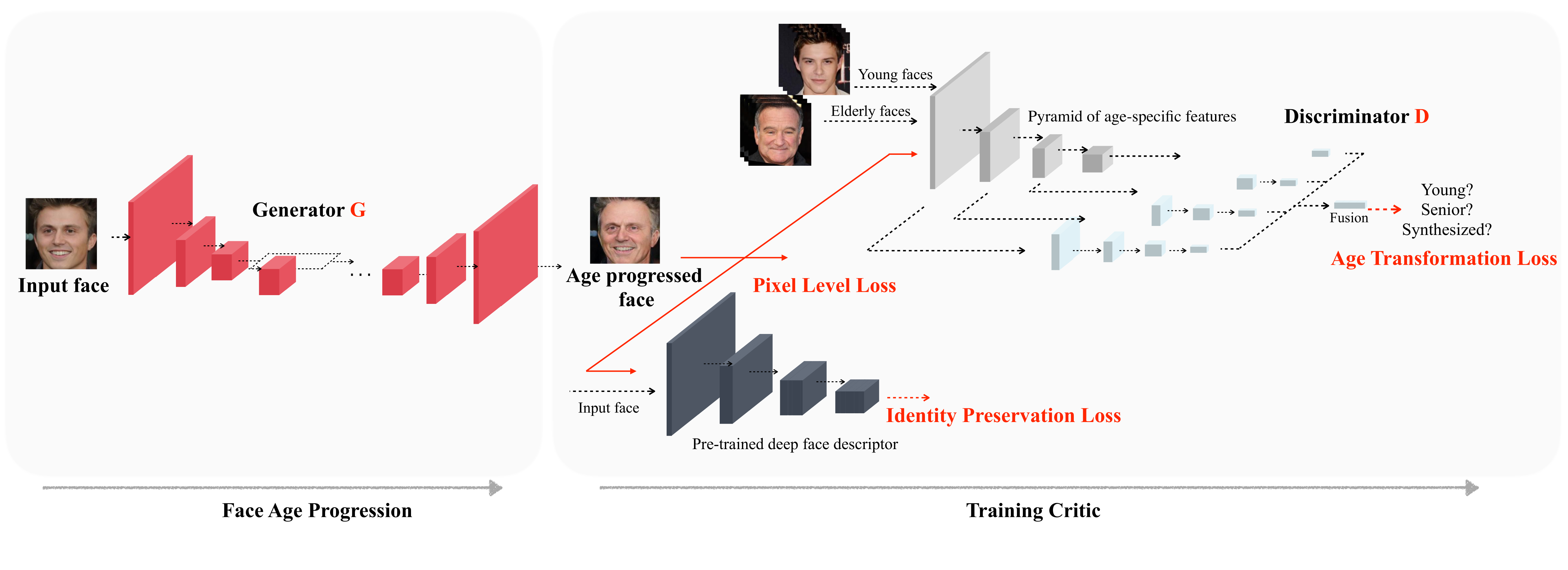}
\caption{Framework of the proposed age progression method. A CNN based generator $G$ learns the age transformation. The training critic incorporates the squared Euclidean loss in the image space, the GAN loss that encourages generated faces to be indistinguishable from the training elderly faces in terms of age, and the identity preservation loss minimizing the input-output distance in a high-level feature representation which embeds the personalized characteristics.}
\vspace{-0.25cm}
\end{figure*}

The majority of the subsequent approaches were data-driven, which did not rely much on the biological prior knowledge, and the aging patterns were learned from the training faces. Wang \emph{et al.} \cite{Wang:tensorAging} built the mapping between corresponding down-sampled and high-resolution faces in a tensor space, and aging details were added on the latter. Kemelmacher-Shlizerman \emph{et al.} \cite{Kemelmacher:aging} presented a prototype based method, and further took the illumination factor into account. Yang \emph{et al.} \cite{Yang:faceAging} first settled the multi-attribute decomposition problem, and progression was achieved by transforming only the age component to a target age group. These methods did improve the results, however ghosting artifacts frequently appeared in the synthesized faces. 

More recently, the deep generative networks have been attempted. In \cite{Wang:Recurrent}, Wang \emph{et al.} transformed faces across different ages smoothly by modeling the intermediate transition states in an RNN model. But multiple face images of various ages of each subject were required at the training stage, and the exact age label of the probe face was needed during test, thus greatly limiting its flexibility. Under the framework of conditional adversarial autoencoder \cite{zhang2017age}, facial muscle sagging caused by aging was simulated, whereas only rough wrinkles were rendered mainly due to the insufficient representation ability of the training discriminator. With the Temporal Non-Volume Preserving (TNVP) aging approach \cite{nhan2017temporal}, the short-term age progression was accomplished by mapping the data densities of two consecutive age groups with ResNet blocks \cite{he2016deepresidual}, and the long-term aging synthesis was finally reached by a chaining of short-term stages. Its major weakness, however, was that it merely considered the probability distribution of a set of faces without any individuality information. As a result, the synthesized faces in a complete aging sequence varied a lot in color, expression, and even identity.

Our study also makes use of the image generation ability of GAN, and presents a different but effective method, where the age-related GAN loss is adopted for age transformation, the individual-dependent critic is used to keep the identity cue stable, and a multi-pathway discriminator is applied to refine aging detail generation. This solution is more powerful in dealing with the core issues of age progression, \textit{i.e.} age accuracy and identity preservation.

\section{Method}

\subsection{Overview}
\vspace{-0.13cm}
A classic GAN contains a generator $G$ and a discriminator $D$, which are iteratively trained via an adversarial process. The generative function $G$ tries to capture the underlying data density and confuse the discriminative function $D$, while the optimization procedure of $D$ aims to achieve the distinguishability and distinguish the natural face images from the fake ones generated by $G$. Both $G$ and $D$ can be approximated by neural networks, \textit{e.g.}, Multi-Layer Perceptron (MLP). The risk function of optimizing this minimax two-player game can be written as:

\begin{scriptsize}
\begin{equation}
\mathcal{V}(D, G) = \min \limits_{G} \max \limits_{D}  \mathbb{E}_{x\sim P_{data} (x)} {\rm log}[D(x)] + \mathbb{E}_{z\sim P_{z} (z)} {\rm log} [1-D(G(z))] 
\end{equation}
\end{scriptsize}\noindent where $z$ is a noise sample from a prior probability distribution $P_{z}$, and $x$ denotes a real face image following a certain distribution $P_{data}$. On convergence, the distribution of the synthesized images $P_{g}$ is equivalent to $P_{data}$. 

Recently, more emphasis has been given to the conditional GANs (cGANs) where the generative model $G$ approximates the dependency of the pre-images (or controlled attributes) and their corresponding targets. cGANs have shown promising results in video prediction \cite{mathieu2015deep}, text to image synthesis \cite{reed2016generative}, image-to-image translation \cite{Phillip2016image2image}\cite{zhu2017unpaired}, \emph{etc}. In our case, the CNN based generator takes young faces as inputs, and learns a mapping to a domain corresponding to elderly faces. To achieve aging effects while simultaneously maintaining person-specific information, a compound critic is exploited, which incorporates the traditional squared Euclidean loss in the image space, the GAN loss that encourages generated faces to be indistinguishable from the training elderly faces in terms of age, and the identity loss minimizing the input-output distance in a high-level feature representation which embeds the personalized characteristics. See Fig. 2 for an overview.

\subsection{Generator}

Synthesizing age progressed faces only requires a forward pass through $G$. The generative network is a combination of encoder and decoder. With the input young face, it first exploits three strided convolutional layers to encode it to a latent space, capturing the facial properties that tend to be stable w.r.t. the elapsed time, followed by four residual blocks \cite{he2016deepresidual} modeling the common structure shared by the input and output faces, similar to the settings in \cite{johnson2016perceptual}. Age transformation to a target image space is finally achieved by three fractionally-strided convolutional layers, yielding the age progression result conditioned on the given young face. Rather than using the max-pooling and upsampling layers to calculate the feature maps, we employ the $3 \times 3$ convolution kernels with a stride of 2, ensuring that every pixel contributes and the adjacent pixels transform in a synergistic manner. All the convolutional layers are followed by Instance Normalization and ReLU non-linearity activation. Paddings are added to the layers to make the input and output have exactly the same size. The architecture of $G$ is shown in the \textbf{supplementary material}.

\subsection{Discriminator}
The system critic incorporates the prior knowledge of the data density of the faces from the target age cluster, and a discriminative network $D$ is thus introduced, which outputs a scalar $D(x)$ representing the probability that $x$ comes from the data. The distribution of the generated faces $P_{g}$ (we denote the distribution of young faces as $x\sim P_{young}$, then $G(x)\sim P_{g}$) is supposed to be equivalent to the distribution $P_{old}$ when optimality is reached. Supposing that we follow the classic GAN \cite{gan}, which uses a binary cross entropy classification, the process of training $D$ amounts to minimizing the loss:

\begin{scriptsize}
\begin{equation}
\mathcal{L}_{GAN\_D} =  - \mathbb{E}_{x \in P_{young} (x)} {\rm log}[1 - D(G(x))] - \mathbb{E}_{x \in P_{old} (x)} {\rm log}[D(x)] 
\end{equation}
\end{scriptsize}

It is always desirable that $G$ and $D$ converge coherently; however, $D$ frequently achieves the distinguishability faster in practice, and feeds back vanishing gradients for $G$ to learn, since the JS divergence is locally saturated. Recent studies, \emph{i.e.} the Wasserstein GAN \cite{wgan}, the Least Squares GAN \cite{mao2016least}, and the Loss-Sensitive GAN \cite{lsgan}, reveal that the most fundamental issue lies in how exactly the distance between sequences of probability distributions is defined. Here, we use the least squares loss substituting for the negative log likelihood objective, which penalizes the samples depending on how close they are to the decision boundary in a metric space, minimizing the Pearson $\mathcal{X}^{2}$ divergence. Further, to achieve more convincing and vivid age-specific facial details, both the actual young faces and the generated age-progressed faces are fed into $D$ as negative samples while the true elderly images as positive ones. Accordingly, the training process alternately minimizes the following:



\begin{scriptsize}
\begin{equation}
\begin{aligned}
\mathcal{L}_{GAN\_D} = & ~\frac{1}{2} \mathbb{E}_{x\sim P_{old} (x)} [(D_{\omega}(\phi_{age}(x))-1)^{2}] 
\\& + \frac{1}{2} \mathbb{E}_{x\sim P_{young} (x)} [D_{\omega}(\phi_{age}(G(x)))^{2} + D_{\omega}(\phi_{age}(x))^{2}]
\end{aligned}
\end{equation}

\begin{equation}
\mathcal{L}_{GAN\_G} = \mathbb{E}_{x\sim P_{young} (x)} [(D_{\omega}(\phi_{age}(G(x)))-1)^{2}]
\end{equation} 
\end{scriptsize}
\vspace{-0.15cm}

Note, in (3) and (4), a function $\phi_{age}$ bridges $G$ and $D$, which is especially introduced to extract age-related features conveyed by faces, as shown in Fig. 2. Considering that human faces at diverse age groups share a common configuration and similar texture properties, a feature extractor $\phi_{age}$ is thus exploited independently of $D$, which outputs high-level feature representations to make the generated faces more distinguishable from the true elderly faces in terms of age. In particular, $\phi_{age}$ is pre-trained for a multi-label classification task of age estimation with the VGG-16 structure \cite{simonyan2014very}, and after convergence, we remove the fully connected layers and integrate it into the framework. Since natural images exhibit multi-scale characteristics and along the hierarchical architecture, $\phi_{age}$ captures the properties gradually from exact pixel values to high-level age-specific semantic information, this study leverages the intrinsic pyramid hierarchy. The pyramid facial feature representations are jointly estimated by $D$ at multiple scales, handling aging effect generation in a fine-grained way. 

The outputs of the 2nd, 4th, 7th and 10th convolutional layers of $\phi_{age}$ are used. They pass through the pathways of $D$ and finally result in a concatenated 12 $\times$ 3 representation. In $D$, all convolutional layers are followed by Batch Normalization and LeakyReLU activation except the last one in each pathway. The detailed architecture of $D$ can be found in the \textbf{supplementary material}, and the joint estimation on the high-level features is illustrated in Fig. 3.

\begin{figure}[t] 
\centering
\includegraphics[width=3.2in]{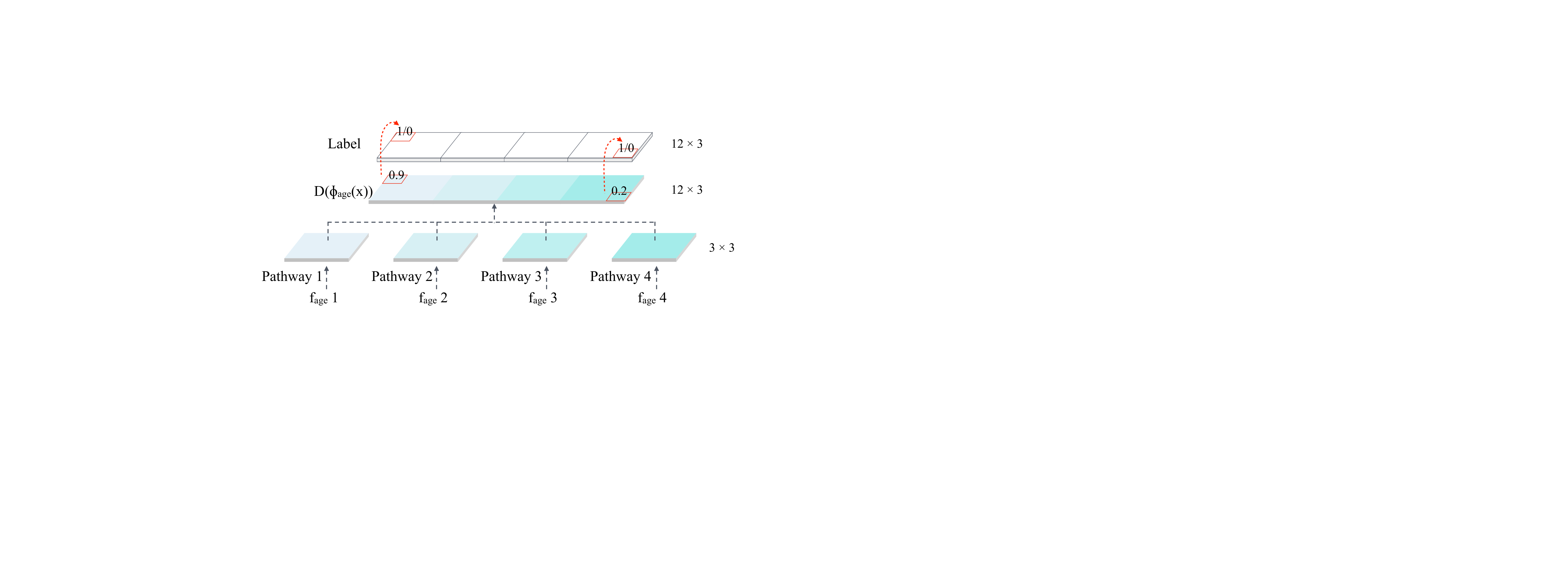}
\caption{The scores of four pathways are finally concatenated and jointly estimated by the discriminator $D$ ($D$ is an estimator rather than a classifier; the $Label$ does not need to be a single scalar).} 
\vspace{-0.15cm}
\end{figure}

\subsection{Identity Preservation}

One core issue of face age progression is keeping the person-dependent properties stable. Therefore, we incorporate the associated constraint by measuring the input-output distance in a proper feature space, which is sensitive to the identity change while relatively robust to other variations. Specifically, the network of \textit{deep face descriptor} \cite{Deepface} is utilized, denoted as $\phi_{id}$, to encode the personalized information and further define the identity loss function. $\phi_{id}$ is trained with a large face dataset containing millions of face images from thousands of individuals\footnote{The face images are collected via the Google Image Search using the names of 5K celebrities, purified by automatic and manual filtering.}. It is originally bootstrapped by recognizing $N = 2,622$ unique individuals; and then the last classification layer is removed and $\phi_{id}(x)$ is tuned to improve the capability of verification in the Euclidean space using a triplet-loss training scheme. In our case, $\phi_{id}$ is clipped to have 10 convolutional layers, and the identity loss is then formulated as:

\begin{scriptsize}
\begin{equation}
\mathcal{L}_{identity} = \mathbb{E}_{x\in P_{young} (x)} d(\phi_{id}(x),\phi_{id}(G(x)))
\end{equation}
\end{scriptsize}\noindent where $d$ is the squared Euclidean distance between feature representations. For more implementation details of \textit{deep face descriptor}, please refer to \cite{Deepface}.

\subsection{Objective}

Besides the specially designed age-related GAN critic and the identity permanence penalty, a pixel-wise L2 loss in the image space is also adopted for further bridging the input-output gap, \textit{e.g.,} the color aberration, which is formulated as: 

\begin{scriptsize}
\begin{equation}
\mathcal{L}_{pixel} = \frac{1}{W \times H \times C}\|G(x) - x\|_{2}^{2}
\end{equation}
\end{scriptsize}\noindent where $x$ denotes the input face and $W$, $H$, and $C$ correspond to the image shape. 

Finally, the system training loss can be written as:

\begin{scriptsize}
\begin{equation}
\mathcal{L}_{G} = \lambda_{a}\mathcal{L}_{GAN\_G} + \lambda_{p}\mathcal{L}_{pixel} +  \lambda_{i}\mathcal{L}_{identity}
\end{equation}

\begin{equation}
\mathcal{L}_{D} = \mathcal{L}_{GAN\_D}
\end{equation}
\end{scriptsize}
\indent We train $G$ and $D$ alternately until optimality, and finally $G$ learns the desired age transformation and $D$ becomes a reliable estimator.

\section{Experimental Results}

\subsection{Data Collection}
\vspace{-0.1cm}
The sources of face images for training GANs are the MORPH mugshot dataset \cite{morph} with standardized imaging and the Cross-Age Celebrity Dataset (CACD) \cite{2015cacd} involving PIE variations.

An extension of the \textbf{MORPH} aging database \cite{morph} contains 52,099 color images with near-frontal pose, neutral expression, and uniform illumination (some minor pose and expression variations are indeed present). The subject age ranges from 16 to 77 years old, with the average age being approximately 33. The longitudinal age span of a subject varies from 46 days to 33 years. \textbf{CACD} is a public dataset \cite{2015cacd} collected via the Google Image Search, containing 163,446 face images of 2,000 celebrities across 10 years, with age ranging from 14 to 62. The dataset has the largest number of images with age changes, showing variations in pose, illumination, expression, \emph{etc.}, with less controlled acquisition than MORPH.
We mainly use MORPH and CACD  for training and validation. FG-NET \cite{fgnet} is also adopted for testing to make a fair comparison with prior work, which is popular in aging analysis but only contains 1,002 images from 82 individuals. More properties of these databases can be found in the \textbf{supplementary material}.

\subsection{Implementation Details}
\vspace{-0.1cm}
Prior to feeding the images into the networks, the faces are aligned using the eye locations provided by the dataset itself (CACD) or detected by the online face recognition API of Face++ \cite{facepp} (MORPH). Excluding those images undetected in MORPH, 163,446 and 51,699 face images from the two datasets are finally adopted, respectively, and they are cropped to 224 $\times$ 224 pixels. Due to the fact that the number of faces older than 60 years old is quite limited in both databases and neither contains images of children, we only consider adult aging. We follow the time span of 10 years for each age cluster as reported in many previous studies \cite{Yang:faceAging}\cite{Suo:Compositional}\cite{zhang2017age}\cite{Wang:Recurrent}\cite{nhan2017temporal}, and apply age progression on the faces below 30 years old, synthesizing a sequence of age-progressed renderings when they are in their 30s, 40s, and 50s. Therefore, there are three separate training sessions for different target age groups.

\begin{figure*}[t]
\centering
\includegraphics[width=1\textwidth]{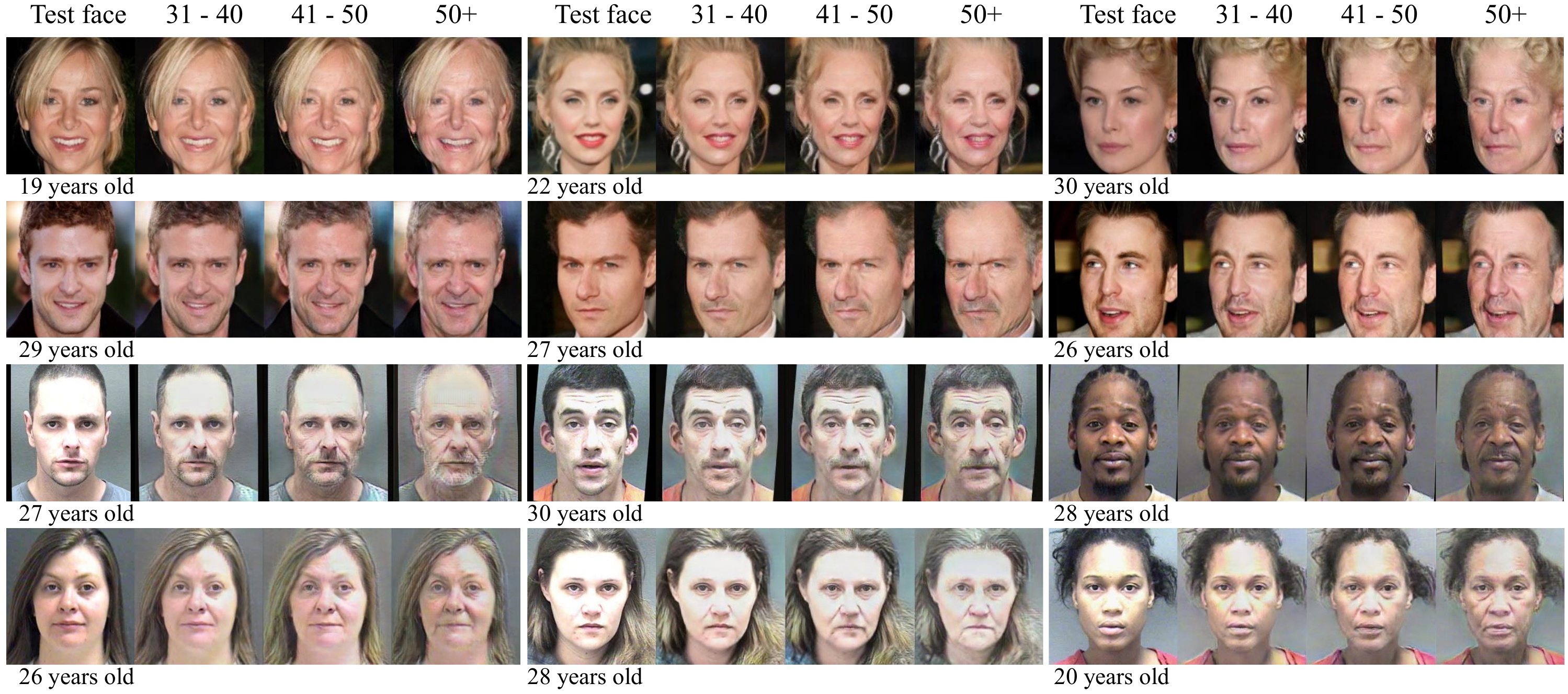}
\caption{Aging effects obtained on the CACD (the first two rows) and MORPH (the last two rows) databases for 12 different subjects. The first image in each panel is the original face image and the subsequent 3 images are the age progressed visualizations for that subject in the [31- 40], [41-50] and 50+ age clusters.}
\vspace{-0.15cm}
\end{figure*}

The architectures of the networks $G$ and $D$ are shown in the \textbf{supplementary material}. For MORPH, the trade-off parameters $\lambda_{p}$, $\lambda_{a}$, and  $\lambda_{i}$ are empirically set to 0.10, 300.00 and 0.005, respectively; and they are set to 0.20, 750.00 and 0.005 for CACD. At the training stage, we use Adam with the learning rate of $1\times {10}^{-4}$ and the weight decay factor of $0.5$ for every $2,000$ iterations. We (i) update the discriminator at every iteration, (ii) use the age-related and identity-related critics at every generator iteration, and (iii) employ the pixel-level critic for every 5 generator iterations. The networks are trained with a batch size of 8 for $50,000$ iterations in total, which takes around 8 hours on a GTX 1080Ti GPU. 

\vspace{-0.05cm}
\subsection{Performance Comparison}
\subsubsection{\textbf{Experiment I: Age Progression}}
\vspace{-0.06cm}
Five-fold cross validation is conducted. On CACD, each fold contains 400 individuals with nearly 10,079,  ~8,635,  ~7,964, and  ~6,011 face images from the four age clusters of [14-30], [31-40], [41-50], and [51-60], respectively; while on MORPH, each fold consists of nearly 2,586 subjects with 4,467, ~3,030, ~2,205, and 639 faces from the four age groups. For each run, four folds are utilized for training, and the remainder for evaluation. Examples of age progression results are depicted in Fig. 4. As we can see, although the examples cover a wide range of population in terms of race, gender, pose, makeup and expression, visually plausible and convincing aging effects are achieved.

\begin{figure}[h]
\centering
\includegraphics[width=3.4in]{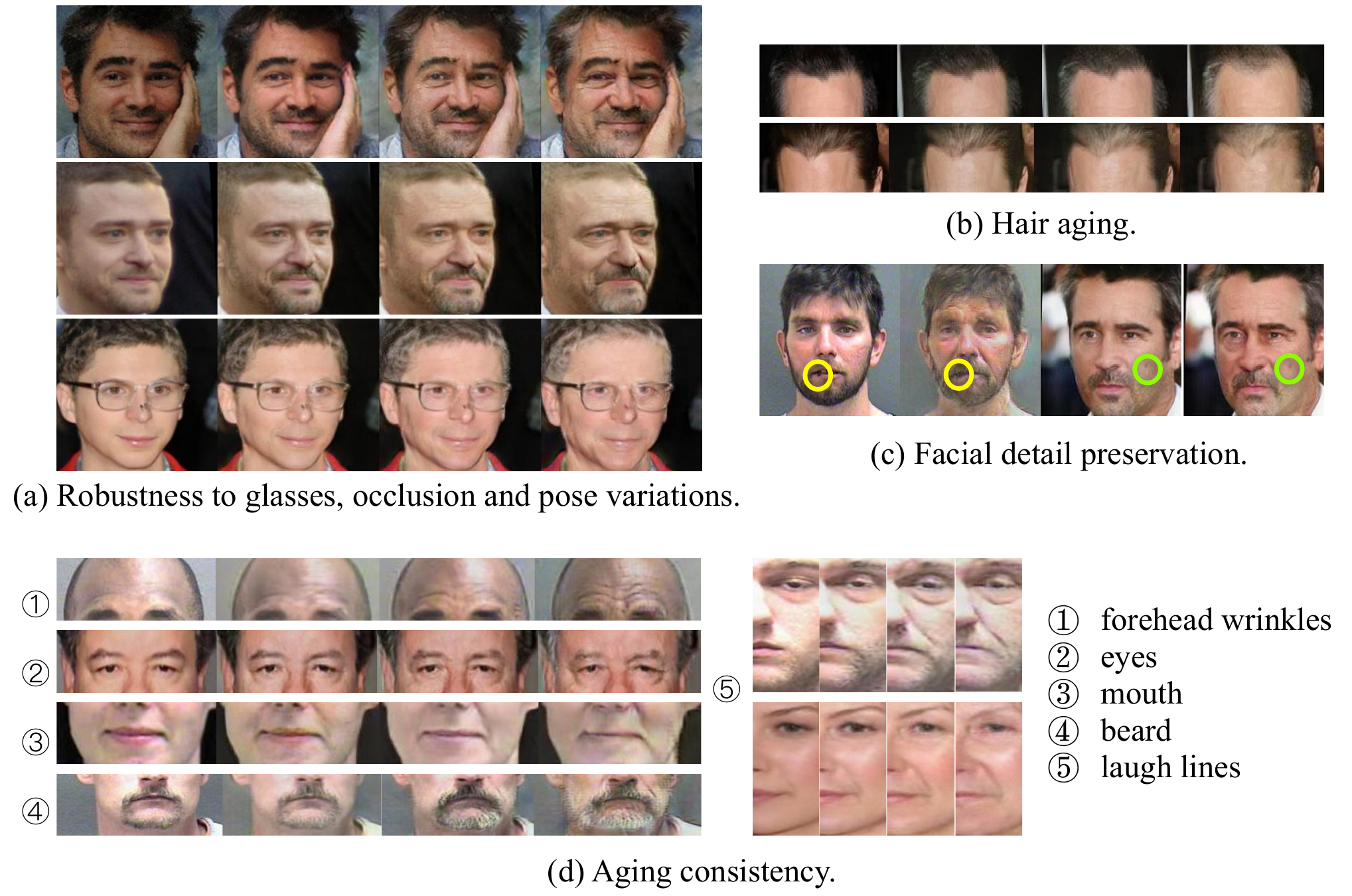}
\caption{Illustration of visual fidelity (zoom in for a better view).}
\vspace{-0.2cm}
\end{figure}

\vspace{-0.3cm}
\subsubsection{\textbf{Experiment II: Aging Model Evaluation}}
\vspace{-0.1cm}
We acknowledge that face age progression is supposed to aesthetically predict the future appearance of the individual, beyond the emerging wrinkles and identity preservation, therefore in this experiment a more comprehensive evaluation of the age progression results is provided with both the visual and quantitative analysis. 

\textbf{Experiment II-A: Visual Fidelity:}
Fig. 5 (a) displays example face images with glasses, occlusions, and pose variations. The age-progressed faces are still photorealistic and true to the original inputs; whereas the previous prototyping based methods \cite{Suo:Compositional}\cite{wang2006age} are inherently inadequate for such circumstances, and the parametric aging models \cite{shu2015personalized}\cite{Suo:Concatenational} may also lead to ghosting artifacts. In Fig. 5 (b), some examples of hair aging are demonstrated. As far as we know, almost all aging approaches proposed in the literature \cite{Yang:faceAging}\cite{shu2015personalized}\cite{Kemelmacher:aging}\cite{Wang:Recurrent}\cite{zhang2017age}\cite{liu2017aging} focus on cropped faces without considering hair aging, mainly because hair is not as structured as the face area. Further, hair is diverse in texture, shape, and color, thus difficult to model. Nevertheless, the proposed method takes the whole face as input, and, as expected, the hair grows wispy and thin in aging simulation. Fig. 5 (c) confirms the capability of preserving the necessary facial details during aging, and Fig. 5 (d) shows the smoothness and consistency of the aging changes, where the lips become thinner, the under-eye bags become more and more obvious, and wrinkles are deeper.

\textbf{Experiment II-B:  Aging Accuracy:} 
Along with face aging, the estimated age is supposed to increase. Correspondingly, objective age estimation is conducted to measure the aging accuracy. We apply the online face analysis tool of Face++ \cite{facepp} to every synthesized face. Excluding those undetected, the age-progressed faces of 22,318 test samples in the MORPH dataset are investigated (average of 4,464 test faces in each run under 5-fold cross validation). Table 1 shows the results. The mean values are 42.84, 50.78, and 59.91 years old for the 3 age clusters, respectively. Ideally, they would be observed in the age range of [31-40] [41-50], and [51-60]. Admittedly, the lifestyle factors may accelerate or slow down the aging rates for the individuals, leading to deviations in the estimated age from the actual age, but the overall trends should be relatively robust. Due to such intrinsic ambiguities, objective age estimations are further conducted on all the faces in the dataset as benchmark. In Table 1 and Fig. 6(a) and 6(c), it can be seen that the estimated ages of the synthesized faces are well matched with those of the real images, and increase steadily with the elapsed time, clearly validating our method.

On CACD, the aging synthesis results of $50,222$ young faces are used in this evaluation (average of 10,044 test faces in each run). Even though the age distributions of different clusters do not have a good separation as in MORPH, it still suggests that the proposed age progression method has indeed captured the data density of the given subset of faces in terms of age. See Table 1 and Figs. 6(b) and 6(d) for detailed results.

\begin{table*}[t] \scriptsize
\centering
\caption{Objective age estimation results (in years) on MORPH and CACD}
~\\
\label{faceverification}
\begin{threeparttable}
\begin{tabular}{ccccccccc}
 \multicolumn{4}{c}{  MORPH } &  &\multicolumn{4}{c}{ CACD}  \\
  \textbf{Age Cluster 0}  &\textbf{Age Cluster 1}  & \textbf{Age Cluster  2} &  \textbf{Age Cluster  3} &  &\textbf{Age Cluster 0}  & \textbf{Age Cluster  1}  & \textbf{Age Cluster  2} &  \textbf{Age Cluster  3}  \\  
\midrule
 \multicolumn{4}{c}{ Synthesized faces\tnote{*}} &  &\multicolumn{4}{c}{Synthesized faces\tnote{*}}  \\
\cline{2-4}
\cline{7-9}
\specialrule{0em}{1pt}{1pt}
\text{--} &  42.84 $\pm$ 8.03 & 50.78 $\pm$ 9.01  & 59.91 $\pm$ 8.95  & &  \text{--} & 44.29 $\pm$ 8.51 & 48.34 $\pm$ 8.32 & 52.02 $\pm$ 9.21 \\
\text{--} &  42.84 $\pm$ 0.40 & 50.78 $\pm$ 0.36  &  59.91 $\pm$ 0.47 & &\text{--} &  44.29 $\pm$ 0.53   & 48.34 $\pm$ 0.35 & 52.02 $\pm$ 0.19\\
 \multicolumn{4}{c}{Natural faces}  &  &\multicolumn{4}{c}{Natural faces}  \\
\cline{1-4}
\cline{6-9}
\specialrule{0em}{1pt}{1pt}
 32.57 $\pm$ 7.95 & 42.46 $\pm$ 8.23 &  51.30 $\pm$ 9.01 & 61.39 $\pm$ 8.56  & &  38.68 $\pm$ 9.50 & 43.59 $\pm$ 9.41 &48.12 $\pm$ 9.52 & 52.59 $\pm$ 10.48\\
\bottomrule
\end{tabular}

\begin{tablenotes}
\footnotesize
\item[*] The standard deviation in the first row is calculated on all the synthesized faces; the standard deviation in the second row is calculated on the mean values of the 5 folds.

\end{tablenotes}
      
\end{threeparttable}
\end{table*}

\begin{table*}[!h] \scriptsize
\centering
\caption{Objective face verification results on (a)~MORPH and (b)~CACD}
~\\
\label{faceverification}
\begin{threeparttable}
\begin{tabular}{cccccccccc}
& & \textbf{Aged face 1}  & \textbf{Aged face 2} &  \textbf{Aged face 3} & ~& & \textbf{Aged face 1}  & \textbf{Aged face 2} &  \textbf{Aged face 3}  \\  
\midrule
~& &\multicolumn{3}{c}{verification confidence\tnote{a}  } & ~& &\multicolumn{3}{c}{verification confidence\tnote{a} }  \\
\cline{3-5}
\cline{8-10} 
\specialrule{0em}{1pt}{1pt}
\multirow{11}{*}{(a)}  & \textbf{Test face} &  94.64 $\pm$ 0.03 & 91.46 $\pm$ 0.08 & 85.87 $\pm$ 0.25  & &\multirow{11}{*}{(b)}  &  94.13$\pm$0.04 &91.96$\pm$0.12 & 88.60$\pm$0.15 \\
& \textbf{Aged face 1}  & \text{--}  & 94.34 $\pm$  0.06 & 89.92 $\pm$  0.30  & & & \text{--}   & 94.88$\pm$0.16 & 92.63$\pm$0.09\\
& \textbf{Aged face 2} & \text{--}  &  \text{--} & 92.23 $\pm$  0.24  & & &  \text{--}   &  \text{--}   & 94.21$\pm$0.24\\
~ & & \multicolumn{3}{c}{verification confidence \tnote{b} }  & ~& &\multicolumn{3}{c}{verification confidence\tnote{b} }  \\
\cline{3-5}
\cline{8-10} 
\specialrule{0em}{1pt}{1pt}
&\textbf{Test face}  &  94.64 $\pm$ 1.06 & 91.46 $\pm$ 3.65 & 85.87 $\pm$ 5.53 & & & 94.13$\pm$1.19 &91.96$\pm$2.26 & 88.60$\pm$4.19\\
&\textbf{Aged face 1}  & \text{--}  & 94.34 $\pm$  1.64 & 89.92 $\pm$  3.49 & &  &\text{--}   & 94.88$\pm$0.87 & 92.63$\pm$2.10 \\
&\textbf{Aged face 2}  & \text{--}  &  \text{--} & 92.23 $\pm$  2.09 & &  &\text{--}   &  \text{--}   & 94.21$\pm$1.25\\
~ & & \multicolumn{3}{c}{verification rate (threshold = 76.5, FAR = 1e - 5)}  & ~ & &\multicolumn{3}{c}{verification rate (threshold = 76.5, FAR = 1e - 5)}  \\
\cline{3-5}
\cline{8-10} 
\specialrule{0em}{1pt}{1pt}
&\textbf{Test face} & 100 $\pm$ 0 \% & 98.91 $\pm$ 0.40 \% &  93.09 $\pm$ 1.31 \%  & & & 99.99 $\pm$ 0.01 \% & 99.91 $\pm$ 0.05 \% & 98.28 $\pm$ 0.33 \%\\
\bottomrule
\end{tabular}
\begin{tablenotes}
\footnotesize
\item[a] The standard deviation is calculated on the mean values of the 5 folds.
\item[b] The standard deviation is calculated on all the synthesized faces. 
\end{tablenotes}     
\end{threeparttable}
\end{table*}

\textbf{Experiment II-C: Identity Preservation:}
Objective face verification with Face++ is carried out to check if the original identity property is well preserved during age progression. For each test face, we perform comparisons between the input image and the corresponding aging simulation results: [test face, aged face 1], [test face, aged face 2], and [test face, aged face 3]; and statistical analyses among the synthesized faces are conducted, {\em i.e.} [aged face 1, aged face 2], [aged face 1, aged face 3], and [aged face 2, aged face 3]. Similar to Experiment II-B, 22,318 young faces in MORPH and their age-progressed renderings are used in this evaluation, leading to a total of $22,318 \times 6$ verifications. As shown in Table 2, the obtained mean verification rates for the 3 age-progressed clusters are 100\%, 98.91\%, and 93.09\%, respectively, and for CACD, there are $50,222\times 6$ verifications, and the mean verification rates are 99.99\%, 99.91\%, and 98.28\%, respectively, which clearly confirm the ability of identity preservation of the proposed method. Additionally, in Table 2 and Fig. 7, face verification performance decreases as the time elapsed between two images increases, which conforms to the physical effect of face aging \cite{best2018longitudinal}, and it may also explain the better performance achieved on CACD compared to MORPH in this evaluation.

\begin{figure}[t] 
\centering
\includegraphics[width=3.2in]{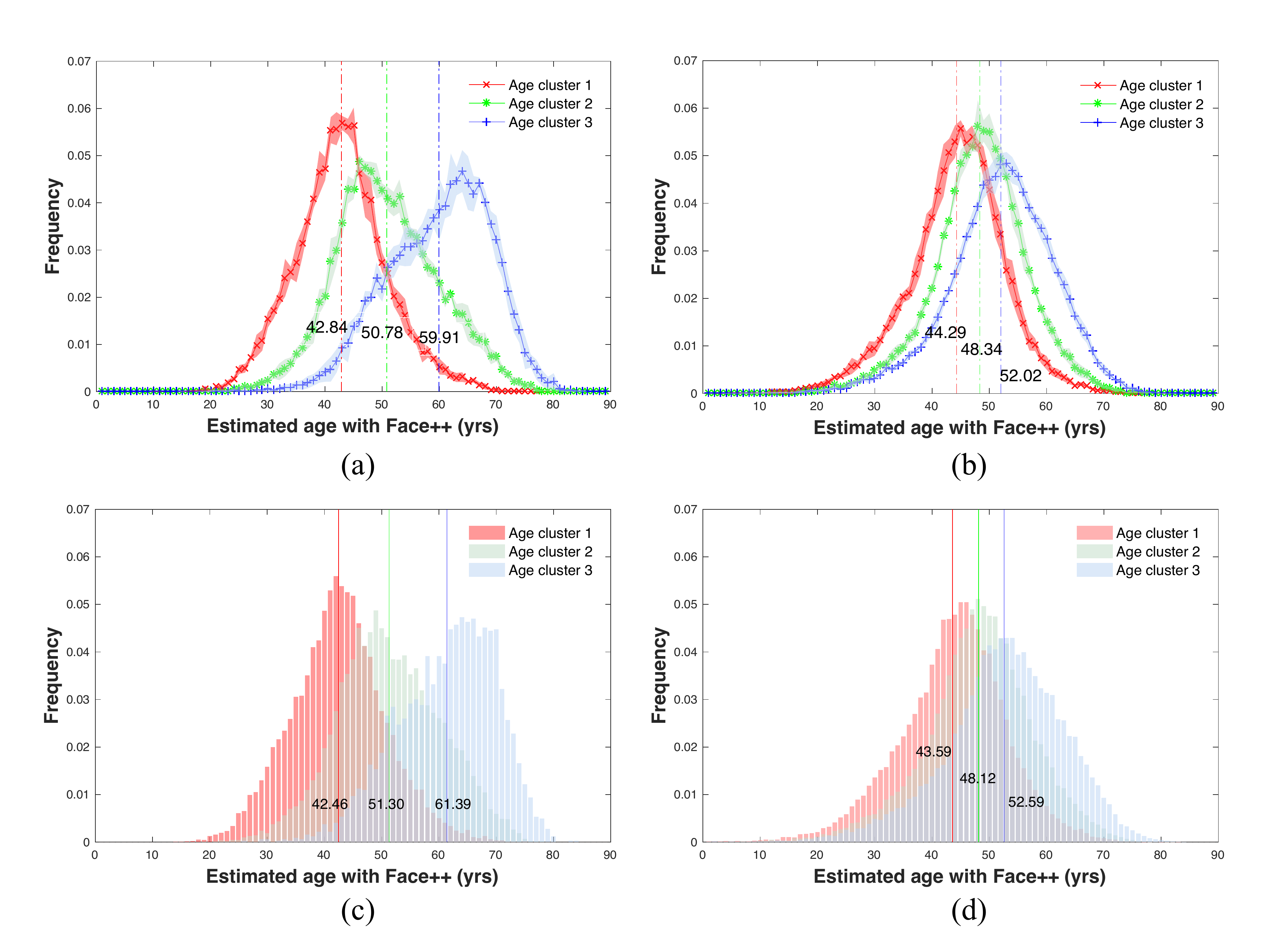}
\caption{Distributions of the estimated ages obtained by Face++. (a) MORPH, synthesized faces; (b) CACD, synthesized faces; (c) MORPH, actual faces; and (d) CACD, actual faces.} 
\vspace{-0.2cm}
\end{figure}

\begin{figure}[h] 
\centering
\includegraphics[width=3.3in]{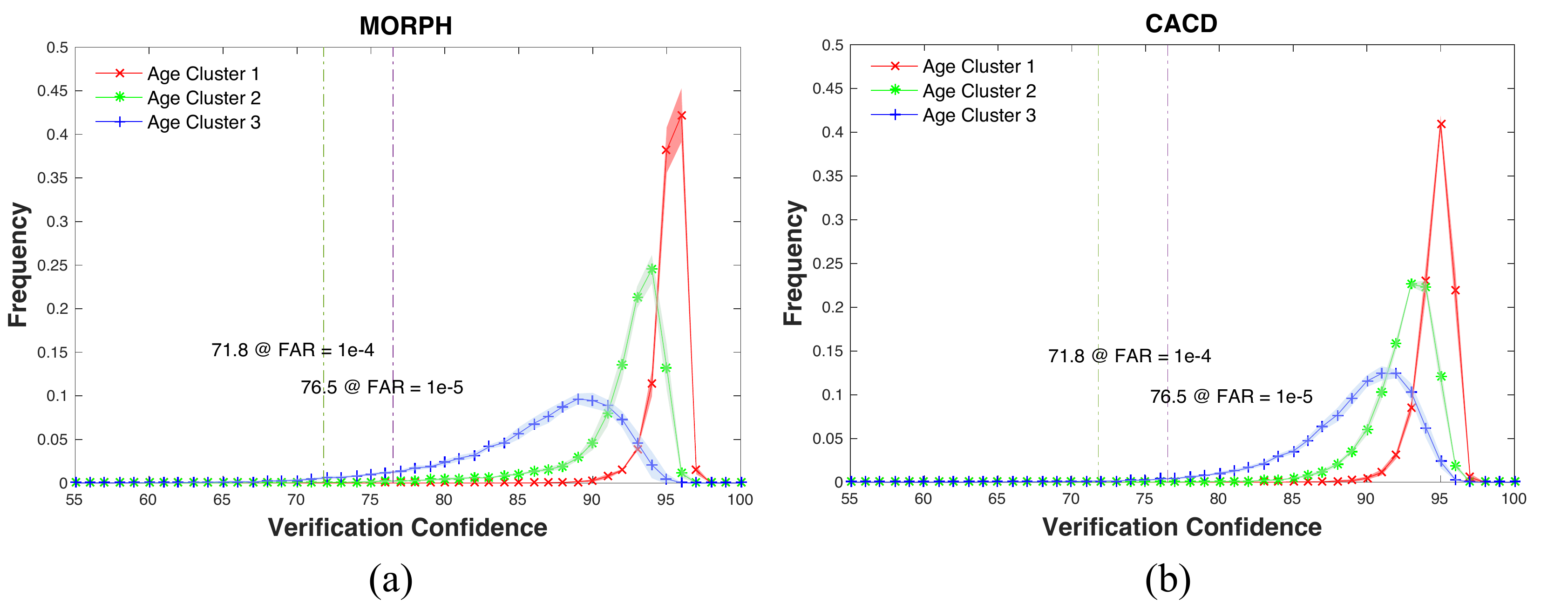}
\caption{Distributions of the face verification confidence on (a) MORPH and (b) CACD.} 
\vspace{-0.2cm}
\end{figure}

\begin{table*}[t] \scriptsize
\centering
\caption{Quantitative evaluation results using one-pathway discriminator on (a) MORPH and (b) CACD }
~\\
\label{faceverification}
\begin{tabular}{cccccccccc}
\toprule
& & \textbf{Aged face 1}  & \textbf{Aged face 2} &  \textbf{Aged face 3} & ~& & \textbf{Aged face 1}  & \textbf{Aged face 2} &  \textbf{Aged face 3}  \\  
\cline{3-5}
\cline{8-10}
\specialrule{0em}{1pt}{1pt}
\multirow{2}{*}{(a)}  & \textbf{Estimated age (yrs old)} &  46.14 $\pm$ 7.79 & 54.99 $\pm$ 7.08 & 62.10 $\pm$ 6.74  & &\multirow{2}{*}{(b)}  &  45.89 $\pm$ 9.85 & 51.44 $\pm$ 9.78 & 54.52 $\pm$ 10.22 \\
\specialrule{0em}{1pt}{1pt}
&\textbf{Verification confidence}  & 93.66 $\pm$ 1.15  & 89.94 $\pm$ 2.59 & 84.51 $\pm$  4.36 & &  & 92.98 $\pm$ 1.76   & 87.55 $\pm$ 4.62 & 84.61 $\pm$ 5.83 \\

\bottomrule
\end{tabular}
\vspace{-0.1cm}
\end{table*}

\vspace{-0.3cm}
\textbf{Experiment II-D: Contribution of Pyramid Architecture:}
One model assumption is that the pyramid structure of the discriminator $D$ advances the generation of the aging effects, making the age-progressed faces more natural. Accordingly, we carry out comparison to the one-pathway discriminator, under which the generated faces are directly fed into the estimator rather than represented as feature pyramid first. The discriminator architecture in the contrast experiment is equivalent to a chaining of the network $\phi_{age}$ and the first pathway in the proposed pyramid $D$.

\begin{figure}[t] 
\centering
\includegraphics[width=3.1in]{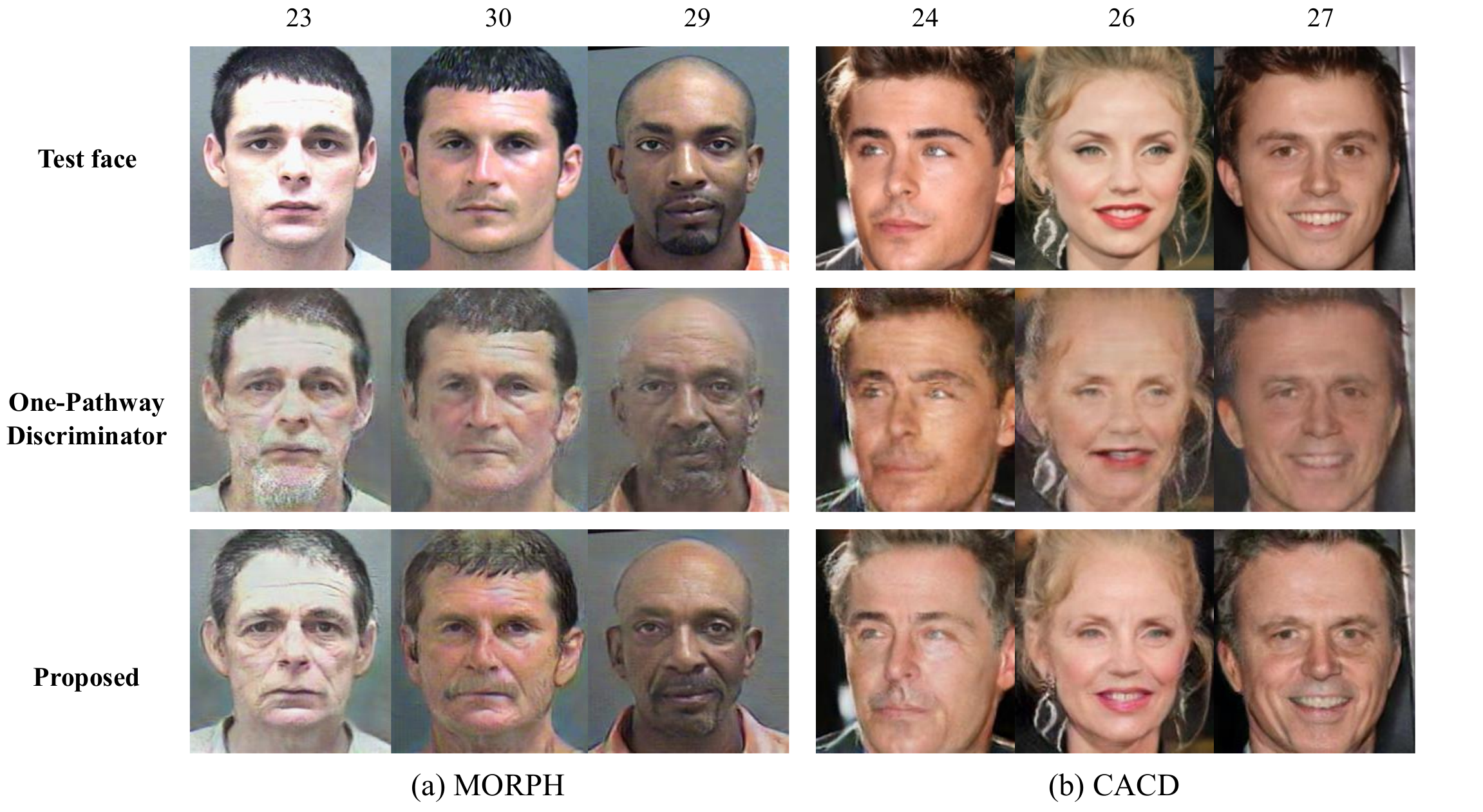}
\caption{Visual comparison to the one-pathway discriminator.}
\vspace{-0.2cm}
\end{figure}

\begin{figure*}[!h] 
\vspace{-0.15cm}
\centering
\includegraphics[width=6.47in]{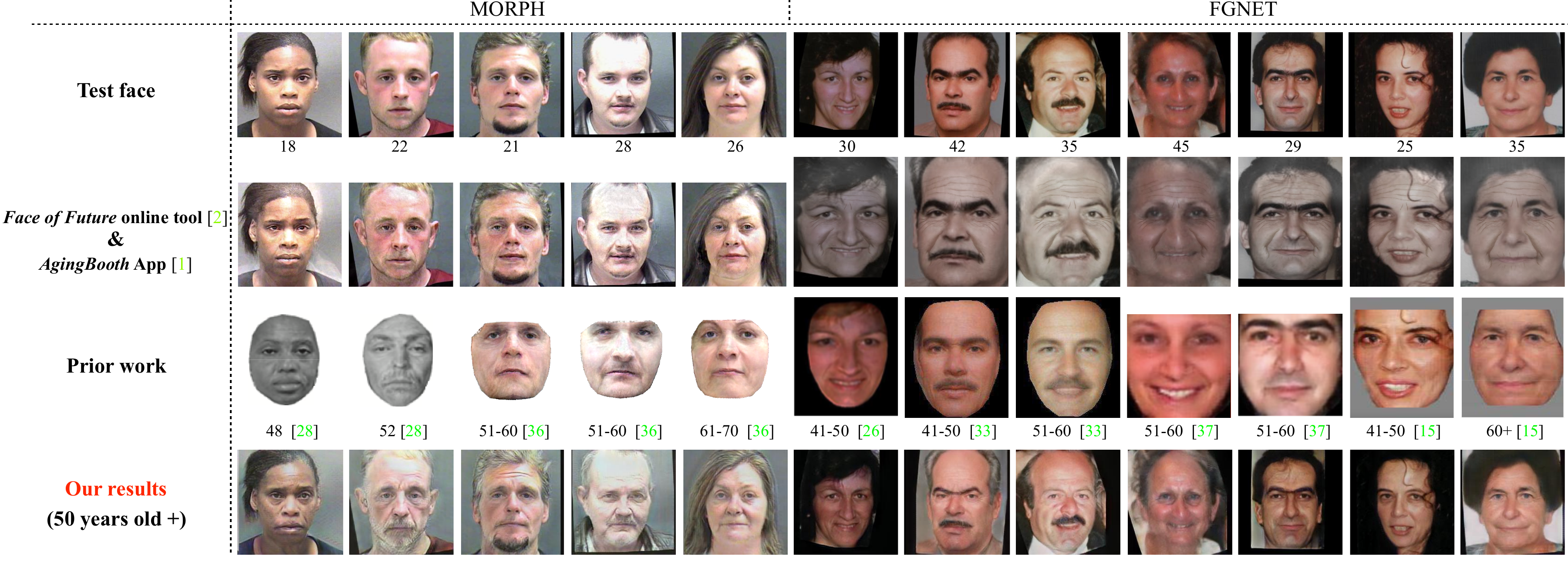}
\caption{Performance comparison with prior work (zoom in for a better view of the aging details).}
\vspace{-0.3cm}
\end{figure*}

Fig. 8 provides a demonstration. Visually, the synthesized aging details of the counterpart are not so evident. To make the comparison more specific and reliable, quantitative evaluations are further conducted with the similar settings as in Experiments II-B and II-C, and the statistical results are shown in Table 3. In the table, the estimated ages achieved on MORPH and CACD are generally higher than the benchmark (shown in Table 1), and the mean absolute errors over the three age clusters are 2.69 and 2.52 years for the two databases, respectively, exhibiting larger deviation than 0.79 and 0.50 years obtained by using the pyramid architecture. It is perhaps because the synthesized wrinkles in this contrast experiment are less clear and the faces look relatively messy. It may also explain the decreased face verification confidence observed in Table 3 in the identity preservation evaluation. Based on both the visual fidelity and the quantitative estimations, we can draw an inference that compared with the pyramid architecture, the one-pathway discriminator, as widely utilized in previous GAN-based frameworks, is lagging behind in regard to modeling the sophisticated aging changes.

\textbf{Experiment II-E: Comparison to Prior Work:}
To compare with prior work, we conduct testing on the FG-NET and MORPH databases with CACD as the training set. These prior studies are \cite{shu2015personalized}\cite{Suo:Concatenational}\cite{Wang:Recurrent}\cite{Yang:faceAging}\cite{nhan2016longitudinal}\cite{zhang2017age}\cite{nhan2017temporal}\cite{park2010age}\cite{liu2017aging}, which signify the state-of-the-art. In addition, one of the most popular mobile aging applications, {\em i.e.} \textit{Agingbooth} \cite{agingbooth}, and the online aging tool \textit{Face of the future} \cite{futureface} are also compared. Fig. 9 displays some example faces. As can be seen, \textit{Face of the future} and \textit{Agingbooth} follow the prototyping-based method, where the identical aging mask is directly applied to all the given faces as most of the aging Apps do. While the concept of such methods is straightforward, the age-progressed faces are not photorealistic. Regarding the published works in the literature, ghosting artifacts are unavoidable for the parametric method \cite{Suo:Concatenational} and the dictionary reconstruction based solutions \cite{Yang:faceAging}\cite{shu2015personalized}. Technological advancements can be observed in the deep generative models \cite{Wang:Recurrent}\cite{zhang2017age}\cite{liu2017aging}, whereas they only focus on the cropped facial area, and the age-progressed faces lack necessary aging details. In a further experiment, we collect 138 paired images of 54 individuals from the published papers, and invite 10 human observers to evaluate which age-progressed face is better in the pairwise comparison. Among the 1,380 votes, 69.78\% favor the proposed method, 20.80\% favor the prior work, and 9.42\% indicate that they are about the same. Besides, the proposed method does not require burdensome preprocessing as previous works do, and it only needs 2 landmarks for pupil alignment. To sum up, we can say that the proposed method outperforms the counterparts.

\section{Conclusions}
\vspace{-0.15cm}
Compared with the previous approaches to face age progression, this study shows a different but more effective solution to its key issues, \textit{i.e.} age transformation accuracy and identity preservation, and proposes a novel GAN based method. This method involves the techniques on face verification and age estimation, and exploits a compound training critic that integrates the simple pixel-level penalty, the age-related GAN loss achieving age transformation, and the individual-dependent critic keeping the identity information stable. For generating detailed signs of aging, a pyramidal discriminator is designed to estimate high-level face representations in a finer way. Extensive experiments are conducted, and both the achieved aging imageries and the quantitative evaluations clearly confirm the effectiveness and robustness of the proposed method.

\section{Acknowledgment}
\vspace{-0.15cm}
This work is partly funded by the National Key Research and Development Plan of China (No. 2016YFC0801002), the National Natural Science Foundation of China (No. 61673033 and No. 61421003), and China Scholarship Council (201606020050).

{\small
\bibliographystyle{ieee}
\bibliography{egbib}
}

\end{document}